\documentclass{article}
\usepackage[preprint]{tmlr}

\usepackage{graphicx}
\usepackage{booktabs}
\usepackage{amsmath,amssymb}
\usepackage{microtype}
\usepackage[hidelinks]{hyperref}  

\newcommand{\auroc}{\textsc{auroc}}

\title{What Predicts Correctness in Text-to-SQL? A Selective-Prediction Study}

\author{Robert Richardson \\
  Affiliation \\
  \texttt{rebelrob10@gmail.com}
}

\begin{document}
\maketitle

\begin{abstract}
Evaluating uncertainty in AI-generated SQL queries requires an estimate of whether it is correct, where
correct means the query executes to the same result as a human-written reference query. We study
which signals provide correctness on hard multi-table text-to-SQL using \auroc{} to compare how well various methods rank correct queries above incorrect ones. Using BIRD and Spider SQL query datasets, the black-box statistical signals such as string, structural, and execution self-consistency, a schema-relevance score,
and query executability all fall between about $0.61$ and $0.68$ \auroc{}, with string
self-consistency the strongest at $0.675$. White-box log-probability is similar ($0.67$). The
signals that move past this ceiling are verification-based, with an LLM judge scoring from $0.72$
(GPT-4o-mini) to $0.78$ (Claude). Judges from different providers make different errors, so a two-provider ensemble reaches $0.82$ \auroc{} with a well-calibrated
probability (expected calibration error $0.03$), and it supports useful abstention frontiers (for
example, answering $27\%$ of questions at $24\%$ selective risk) where self-consistency offers no
valid low-risk subset. The pattern holds across two benchmarks (BIRD and Spider), two generators, and
two judge providers. We also explore whether a verifier can be trained. Fine-tuned verifiers, both encoder and generative, reach about
$0.77$--$0.79$ \auroc{} in-distribution but fall to about $0.66$ on unseen schemas, and scaling to
7B, adding schema diversity, distilling a strong judge's rationales, and cross-benchmark training all
fail to close that gap. Cross-schema transfer appears to track model scale and reasoning rather than
fine-tuning. In practice, correctness uncertainty for text-to-SQL lives in reasoning-based signals, meaning that a
fine-tuned verifier is a good in-domain tool, while a verifier that generalizes across schemas
currently means a large frozen reasoning model.
\end{abstract}

\section{Introduction}
A text-to-SQL system turns a natural-language question into a SQL query that runs against a database.
These systems are increasingly used where a wrong query is costly, so a system needs to know not just
how to answer but when it should decline to. Knowing when to decline requires, for each query the
system produces, an estimate of how likely that query is to be correct. We use \emph{correct} in its
standard benchmark sense: the generated query, when executed against the database, returns the same
result as a human-written reference query for the same question. Predicting correctness so a system
can choose to answer or to abstain is the problem of selective prediction, and it is the focus
of this paper.

Most uncertainty-quantification (UQ) work for text-to-SQL
reads it off statistics of the model's own outputs, without any extra model or labels. The clearest
example is self-consistency: draw several queries for the same question and measure how much
they agree, on the premise that a confident model repeats itself \citep{wang2023selfconsistency}.
Related signals include the entropy of the samples' meanings \citep{kuhn2023semantic,farquhar2024semantic}
and how often a particular query structure recurs. We call these \emph{black-box statistical} signals
because they look only at agreement among samples, not at whether a query's logic actually answers the
question. The alternative is verification: show a separate model the question, the schema, and
the candidate query, and ask it to judge whether the query is correct (also called LLM-as-judge). We
ask a direct question: on hard text-to-SQL, which of these signals actually predicts correctness, and
by how much?

To answer it we score each signal by its ability to rank correct queries above incorrect ones,
measured as the area under the ROC curve (\auroc{}), which is the probability that a randomly chosen correct
query is scored above a randomly chosen incorrect one, where $0.5$ is chance and $1.0$ is a perfect
ranking. We attach paired bootstrap confidence intervals to every comparison, so a difference between
two signals can be read as significant or not. Our contribution is this execution-grounded comparison
of the signals available to a practitioner, together with what it implies for deployment.
\begin{enumerate}
\item A correctness ceiling for black-box statistical UQ (roughly $0.61$--$0.68$ \auroc{}), shared by
string, structural, and execution self-consistency, by a schema-relevance score, and, once compared
against the strongest of these, by white-box log-probability (the model's own probability for the
query it generated).
\item Verification moves past the ceiling, with bootstrapped intervals that exclude zero, across two
benchmarks, two generators, and two judge providers. Judges from different providers make different
errors, and an ensemble of them is the best and best-calibrated correctness signal we obtain ($0.82$
\auroc{}, $0.03$ calibration error).
\item A selective-prediction frontier that the self-consistency baseline cannot match.
\item A study of trained verifiers, fine-tuned on labeled examples. They do well on databases
seen in training but fail to transfer (keep working on databases never seen during training),
which separates an in-domain verifier from a universal one.
\end{enumerate}

Our scope is selective prediction rather than generation. We quantify and calibrate correctness to
decide whether to answer or abstain; we do not iterate to improve the query, in contrast with the
self-correcting agents discussed next.

\section{Related work}
\label{sec:related}
Relevant lines of work that have a similar goal to what we are attempting to do here include self-consistency and sampling-based UQ
\citep{wang2023selfconsistency}, semantic entropy via meaning clustering
\citep{kuhn2023semantic,farquhar2024semantic}, verbalized confidence, conformal and selective
generation \citep{geifman2017selective,angelopoulos2021ltt,angelopoulos2023conformalrisk}, sub-clause
Platt scaling \citep{ramachandran2024calibrating}, and conformal abstention over schema-linking
branch points \citep{somov2025rts}. Schema linking has also been studied with encoder-side graph
models \citep{wang2020ratsql,cao2021lgesql}. LLM-as-judge is widely used for open-ended evaluation.
Here we evaluate it specifically as a calibrated correctness predictor for SQL, against black-box
baselines, and we test whether judging can be trained and whether it transfers. We draw data from the
Spider \citep{yu2018spider} and BIRD \citep{li2023bird} benchmarks, and reference Spider~2.0
\citep{lei2025spider2} for context.

A closely related line builds self-correcting or agentic text-to-SQL systems, in which the model
generates a query and then critiques or executes it and regenerates, often over several rounds
\citep{shinn2023reflexion,chen2023selfdebug,pourreza2023dinsql,wang2024macsql}. That work shares a
component with ours, since the critic in such a loop is an LLM verifier much like the judge we study,
but the objective differs. A self-correcting agent uses the critic to drive iteration toward a more
accurate query and reports end-task accuracy. We do not regenerate, but rather ask how well a signal predicts
correctness so a system can answer or abstain, and we report calibration and risk-coverage rather
than accuracy gains. The two are complementary, because even after the system corrects itself, it still reports a
final query and that query still needs a calibrated estimate of whether to trust it. We check this directly
(Section~\ref{sec:robust}) and find that one round of self-correction barely changes accuracy and produces badly
miscalibrated confidence, so it does not substitute for a calibrated verifier. Typical
self-correction also relies on execution feedback, whereas our verifier reasons over the question,
schema, and SQL without executing the candidate, and we find executability itself uninformative for
correctness.

\section{Setup}
\textit{Benchmark and generation.} We use BIRD \citep{li2023bird}, which has databases with data
dictionaries and external-knowledge ``evidence.'' We generate SQL with GPT-4o-mini ($K=8$ samples
per question, temperature $0.7$, schema and evidence in the prompt) for a fixed slice of 800 questions
spanning 8 databases. The slice was fixed before any analysis and does not depend on model output. It
takes the questions in benchmark order from each of the 8 databases whose files we held locally,
capped per database and at 800 in total. We execute every generated sample against the real database
and label a query ``correct'' when its result set matches the reference query's, which is the
standard execution-correctness criterion for these benchmarks. Throughout, the query we score is the
modal query, the one a question's samples most often produce. Modal-query execution accuracy is
$0.451$, a hard regime with ample headroom for UQ. For breadth we also use a second generator,
GPT-4.1-mini (accuracy $0.522$), and a second benchmark, Spider's \citep{yu2018spider} multi-table dev
queries ($490$ questions over $20$ databases), generated and executed the same way.

\textit{Signals.} The black-box statistical signals are string self-consistency (the size of the
largest cluster of identical queries among the samples), structural self-consistency (clustering by
canonical query structure), execution self-consistency (clustering by result set), a schema-relevance
score (the average embedding similarity between the question and the tables and columns the query
uses), and query executability (whether the query runs and returns rows). The logic-aware signals are
the mean sequence log-probability of the modal query (white-box) and an LLM verifier shown the
question, evidence, schema, and candidate SQL. The verifiers are GPT-4o-mini and GPT-4o, which report
$P(\text{correct})$ from the YES/NO first-token logits, and Claude-Sonnet-4.6, an
independent-provider judge that reports a verbalized $0$--$100$ probability (Anthropic does not expose
token logprobs). All judges see the same input in the same order; the exact prompts are in
Appendix~\ref{app:prompts}.

\textit{Metrics.} We report tie-robust \auroc{} for correctness, with $2{,}000$-sample paired
bootstrap confidence intervals on \auroc{} differences. We also combine signals with a cross-fit
logistic regression, which means the combiner is fit on one split and scored on a disjoint split, so it never
sees its own test data. For calibration we report expected calibration error (ECE), the average gap
between a score read as a probability and the empirical fraction correct at that score
\citep{guo2017calibration}. For abstention we report a risk--coverage frontier, which is the selective risk,
or error rate among answered questions, as a function of coverage, the fraction answered, using both
an empirical frontier and a distribution-free (Bonferroni-over-grid, $\delta=0.1$) certificate
\citep{angelopoulos2021ltt}. All model outputs are cached, and the API experiments are inexpensive.

\section{Results}

\subsection{The black-box correctness ceiling}
\label{sec:ceiling}

Table~\ref{tab:ceiling} shows the results for the black box signals that do not require additional models. It shows that they are not enough to tell correct SQL from incorrect SQL at high rates. The black-box statistical signals occupy a narrow band, roughly $0.61$--$0.68$ \auroc{}, with string self-consistency the strongest at $0.675$. Structural self-consistency and execution self-consistency are weaker, and executability is essentially chance: wrong queries often run successfully and return rows.

\begin{table}[htbp]
\centering
\begin{tabular}{lr}
\hline
Signal (alone) & \auroc{} for correctness \\
\hline
String self-consistency (largest identical-query cluster) & 0.675 \\
Structural self-consistency (canonical-structure cluster) & 0.619 \\
Execution self-consistency (result-set cluster) & 0.613 \\
Log-probability (white-box, mean sequence logprob) & 0.669 \\
Schema-relevance score & 0.553 \\
Executability (runs and returns rows) & 0.500 (chance) \\
\hline
\end{tabular}
\caption{Correctness prediction from single signals on the BIRD slice. Sampling-based and structural signals plateau below the verifier results reported in Section~\ref{sec:verify}.}
\label{tab:ceiling}
\end{table}

White-box log-probability is also not enough in this harder multi-table setting. Its \auroc{} is $0.669$, slightly below string self-consistency, and the paired difference is small and not significant: $-0.006$ with 95\% CI $[-0.028,+0.017]$. This is a useful negative result. In earlier single-table experiments, log-probability helped separate confidently wrong unanimous samples. On BIRD, where the errors involve more complex reasoning, it provides no better correctness signal than sampling agreement. The result suggests a ceiling for signals that do not directly reason about whether the SQL answers the question.

\subsection{Verification moves past the ceiling}
\label{sec:verify}

We next test signals that can inspect the candidate query itself. The verifier is shown the question, evidence, schema, and generated SQL, and asked whether the SQL correctly answers the question. Unlike self-consistency or log-probability, this signal can in principle evaluate the query's logic, deciding whether the aggregation is appropriate, whether the filters match the wording of the question, whether the grouping is correct, and whether the query computes the requested quantity.

Table~\ref{tab:verify} shows that verification is the first signal to move clearly past the black-box ceiling. All three LLM verifiers exceed string self-consistency, and the paired bootstrap intervals exclude zero. GPT-4o-mini gives a moderate gain, while the stronger GPT-4o and Claude judges give substantially larger gains.

\begin{table}[htbp]
\centering
\begin{tabular}{lrr}
\hline
Signal & \auroc{} & Paired $\Delta$ vs string SC (95\% CI) \\
\hline
Log-probability (white-box) & 0.669 & $-0.006\ [-0.028, +0.017]$ (n.s.) \\
LLM verifier (GPT-4o-mini) & 0.724 & $+0.049\ [+0.004, +0.090]$ \\
LLM verifier (GPT-4o) & 0.770 & $+0.095\ [+0.053, +0.138]$ \\
LLM verifier (Claude-Sonnet-4.6) & 0.776 & $+0.101\ [+0.060, +0.141]$ \\
\hline
\end{tabular}
\caption{Verifier scores for execution correctness. Paired differences compare each single signal against string self-consistency.}
\label{tab:verify}
\end{table}

The gain comes from the verifier itself, not from a generic combination with self-consistency. A cross-fit logistic model using GPT-4o plus string self-consistency reaches $0.754$ \auroc{}, which is below GPT-4o alone at $0.770$. In other words, once a strong verifier is available, the sampling agreement signal contributes little. 

\subsection{Robustness: cross-provider judges, a second generator, and self-correction}
\label{sec:robust}

A natural concern is that a verifier from the same model family as the generator might simply share the generator's biases. To test this, we add an independent-provider judge. Claude-Sonnet-4.6, judging GPT-4o-mini's SQL, is the strongest single verifier in this experiment, with \auroc{} $0.776$ compared with GPT-4o's $0.770$. More importantly, the GPT-4o and Claude scores are not redundant as their correlation is only $r=0.43$, indicating that they make different errors.

The pattern is not specific to GPT-4o-mini as the generator. We regenerate the same BIRD slice with GPT-4.1-mini, which is stronger on this task, and judge the resulting modal queries with GPT-4o-mini. The verifier still improves over self-consistency, although the margin is smaller.

\begin{table}[htbp]
\centering
\begin{tabular}{lrrrr}
\hline
Generator & Accuracy & String SC & Verifier & Combined $\Delta$ over SC (95\% CI) \\
\hline
GPT-4o-mini & 0.451 & 0.675 & 0.724 & $+0.059\ [+0.032, +0.087]$ \\
GPT-4.1-mini & 0.522 & 0.641 & 0.677 & $+0.033\ [+0.007, +0.061]$ \\
\hline
\end{tabular}
\caption{The verifier advantage holds for two generators. The margin is smaller when GPT-4o-mini judges the stronger GPT-4.1-mini generator, as expected.}
\label{tab:second-generator}
\end{table}

For both generators, self-consistency remains in the same performance band and the verifier moves beyond it. The smaller gain for GPT-4.1-mini is consistent with the judge being weaker than the generator: subtler errors made by the stronger generator are harder for GPT-4o-mini to detect. This suggests that the judge should be at least as strong as the generator when the goal is universal correctness prediction.

\paragraph{A second benchmark.}
The same qualitative pattern appears on Spider. On Spider's multi-table dev queries ($490$ questions over $20$ databases; modal accuracy $0.620$), string self-consistency again sits at the ceiling ($0.618$ \auroc{}), while a strong verifier clears it: GPT-4o reaches $0.712$ \auroc{}, a paired improvement of $+0.094$ over self-consistency with 95\% CI $[+0.029,+0.155]$. The weaker GPT-4o-mini judge is above the ceiling but not significantly so ($0.669$, $\Delta=+0.051$, CI $[-0.013,+0.114]$). Across two benchmarks, two generators, and two judge providers, the same lesson holds: self-consistency plateaus, and a sufficiently strong verifier moves past it.

\paragraph{A self-correction baseline.}
Self-correcting agents are another way to use a critic, so we also test a simple reflection baseline. Each modal query is fed back to GPT-4o-mini, which reviews it, revises it if needed, and reports a confidence. This changes the query, so it addresses a different problem from selective prediction; nevertheless, it is an important comparison because it asks whether a self-correction loop makes a separate trust signal unnecessary.

It does not. One reflection round improves execution accuracy only marginally, from $0.451$ to $0.468$. More importantly for our purposes, the model's self-reported confidence is badly miscalibrated. It assigns confidence at least $0.9$ to $99\%$ of revised queries, with mean confidence $0.97$, but its correctness signal has only $0.637$ \auroc{} and ECE $0.501$. Thus, a single round of self-correction neither closes the accuracy gap nor produces a calibrated abstention score. Regeneration and selective prediction are complementary: even after a model revises a query, the system still needs to decide whether the final query should be trusted.

\subsection{Can the verifier be trained? In-distribution yes, transfer no}
\label{sec:trained}

The frozen LLM verifier is effective, but it requires an additional model call at inference time. A natural alternative is to train a cheaper verifier. We fine-tune models on examples of the form
\[
(\text{question}, \text{evidence}, \text{schema}, \text{SQL}) \rightarrow \text{correct},
\]
using $6{,}400$ execution-labeled pairs. We then evaluate whether the trained verifier transfers to unseen databases. The transfer evaluation uses leave-one-database-out (LODO): one database is held out, the verifier is trained on the remaining databases, and the held-out database is scored. We repeat this for each database and report macro-averaged transfer \auroc{}.

We compare a feature classifier, a fine-tuned encoder classifier, and LoRA-tuned generative judges at two sizes. Table~\ref{tab:trained} summarizes the result.

\begin{table}[htbp]
\centering
\begin{tabular}{lrr}
\hline
Verifier & In-dist.\ \auroc{} & Transfer (LODO, macro) \\
\hline
Feature classifier (embeddings, logprob, SC) & 0.768 & 0.661 \\
Fine-tuned encoder (ModernBERT-base) & 0.785 & 0.670 \\
Generative judge, Qwen2.5-1.5B (LoRA) & 0.766 & 0.659 \\
Generative judge, Qwen2.5-7B (LoRA) & 0.798 & 0.662 \\
Frozen GPT-4o judge (zero-shot) & --- & 0.710 \\
\hline
\end{tabular}
\caption{Fine-tuned verifiers work in-distribution but transfer poorly to unseen databases. Transfer values are macro-averaged over held-out databases.}
\label{tab:trained}
\end{table}

There are four takeaways. First, fine-tuning works in-distribution. The trained models reach approximately $0.77$--$0.80$ \auroc{}, and the 7B generative judge gives the best in-domain result at $0.798$. For a fixed deployment that owns its schemas and can label examples from those schemas, this is a useful operating point.

Second, the same models transfer poorly. Across architectures and scales, the trained verifiers land near $0.66$--$0.67$ on unseen databases. Scaling the generative judge from 1.5B to 7B improves the in-domain number but does not improve LODO transfer. The frozen GPT-4o judge is the exception: when macro-averaged in the same way, it scores $0.710$, and when pooled over all held-out questions it scores $0.770$. Thus the trained models are useful in-domain, but they do not yet behave like universal correctness verifiers.

Third, the transfer gap is not closed by the additional levers we tried. Adding Spider's schemas to increase training diversity, distilling rationales from a strong judge, scaling the generative verifier, and cross-benchmark training all leave transfer in roughly the same band (Appendix~\ref{app:transfer}). The evidence is consistent with the fine-tuned models relying on schema-specific surface regularities rather than learning a robust cross-schema reasoning procedure.

\begin{figure}[htbp]
\centering
\includegraphics[width=0.48\textwidth]{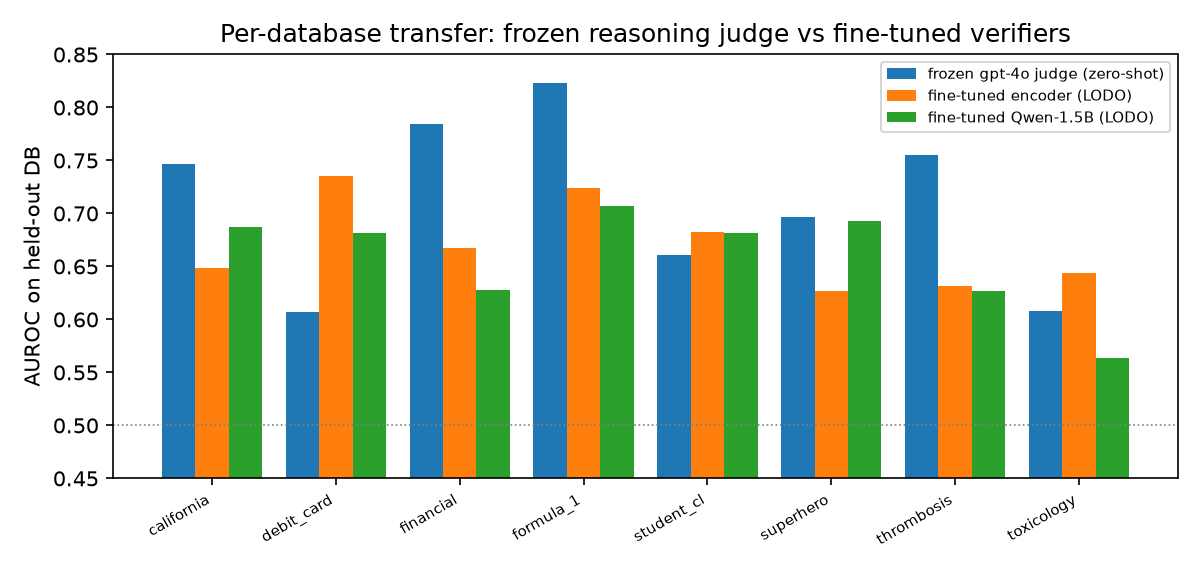}
\caption{Per-database transfer. The frozen GPT-4o judge leads on every held-out schema, above both fine-tuned verifiers. The pattern is consistent with reasoning transferring across schemas while fine-tuning captures schema-specific regularities.}
\label{fig:lodo}
\end{figure}

An input ablation of the frozen judge supports this interpretation. When the judge sees only the question and SQL, it reaches $0.692$ \auroc{}. Adding the schema changes almost nothing ($0.688$), while adding the evidence helps ($0.724$). This suggests that the frozen judge is not mainly looking up schema content; it is reasoning about whether the query answers the question. Per database, the frozen judge leads on every held-out schema (Figure~\ref{fig:lodo}).

Fourth, the in-domain verifier is useful for abstention. A feature classifier with no LLM call at inference gives a risk--coverage frontier competitive with the frozen GPT-4o judge and far above self-consistency. At a $20\%$ risk target it answers $21\%$ of questions, compared with $9\%$ for the frozen judge and $1\%$ for self-consistency; the corresponding areas under the risk--coverage curve are $0.340$, $0.355$, and $0.413$ (lower is better). Thus, for a fixed deployment, a trained verifier is not only accurate but also the cheaper route to a useful abstention policy. For open, cross-schema use, the frozen reasoning judge remains stronger.

\subsection{Calibration and selective prediction}
\label{sec:cal}

For selective prediction, a score must do more than rank correct queries above incorrect ones; it must also support a usable threshold. We therefore examine calibration and risk--coverage. Raw self-consistency and raw verifier probabilities are over-confident, but the cross-fit two-provider ensemble is well-calibrated and can be used directly as an estimate of $P(\text{correct})$.

As stated in Section \ref{sec:robust}, the GPT-4o and Claude scores are not redundant as their correlation is only $r=0.43$, indicating that they make different errors. This difference is useful. A two-provider ensemble of GPT-4o and Claude reaches $0.822$ \auroc{}, improving over GPT-4o alone by $+0.052$ with 95\% CI $[+0.032,+0.073]$. It is the strongest correctness signal we obtain, and it is also well-calibrated after cross-fit logistic calibration (ECE $0.031$; Section~\ref{sec:cal}). The result argues against a self-agreement artifact. The best correctness signal is not merely a larger judge from the same source, but a combination of independent reasoning judgments. The OpenAI judges use YES/NO first-token logits, while Claude reports a verbalized probability; to check that the difference is not just an elicitation artifact, we also evaluate a method-matched OpenAI verbal judge, which still clears the black-box ceiling at $0.709$ \auroc{} (Appendix~\ref{app:prompts}).

\begin{table}[htbp]
\centering
\begin{tabular}{lrr}
\hline
Score & \auroc{} & ECE \\
\hline
String self-consistency & 0.675 & 0.212 \\
Verifier (GPT-4o, raw $P$) & 0.770 & 0.319 \\
Verifier (Claude, raw $P$) & 0.776 & 0.210 \\
Two-provider ensemble (cross-fit) & 0.822 & 0.031 \\
\hline
\end{tabular}
\caption{Calibration and ranking of the main correctness scores. The cross-fit ensemble improves both discrimination and calibration.}
\label{tab:calibration}
\end{table}

We next use the scores for abstention. For each target selective risk, we split the questions in half: the threshold is chosen on the calibration half, and coverage and risk are reported on the disjoint test half. Thus, no threshold is evaluated on the same data that selected it. Table~\ref{tab:selective} reports the resulting empirical frontier.

\begin{table}[htbp]
\centering
\begin{tabular}{lll}
\hline
Target risk & Self-consistency & All signals \\
\hline
0.20 & none (no valid subset) & answers 9\% at 14\% risk \\
0.30 & none & answers 27\% at 24\% risk \\
0.40 & none & answers 58\% at 38\% risk \\
\hline
\end{tabular}
\caption{Empirical selective-prediction frontier. Thresholds are chosen on a calibration half and evaluated on a disjoint test half.}
\label{tab:selective}
\end{table}

Self-consistency cannot form a low-risk subset at any threshold in this split: its most confident region is still too error-prone. The logic-aware ensemble, by contrast, yields useful operating points. At a target risk of $0.30$, for example, it answers $27\%$ of questions with held-out selective risk $0.24$.

We also compute a distribution-free Bonferroni-over-grid certificate. This certificate is conservative in the present regime because the base generator accuracy is only $0.45$. It fires only at the loosest target, certifying the all-signals score at $\alpha=0.40$ with $16\%$ coverage and $22\%$ risk. We therefore treat the empirical frontier as the practical result and the certificate as a conservative lower bound. A tighter certificate would likely require a stronger base generator or more calibration data, rather than a different confidence score.

\begin{figure}[htbp]
\centering
\includegraphics[width=0.48\textwidth]{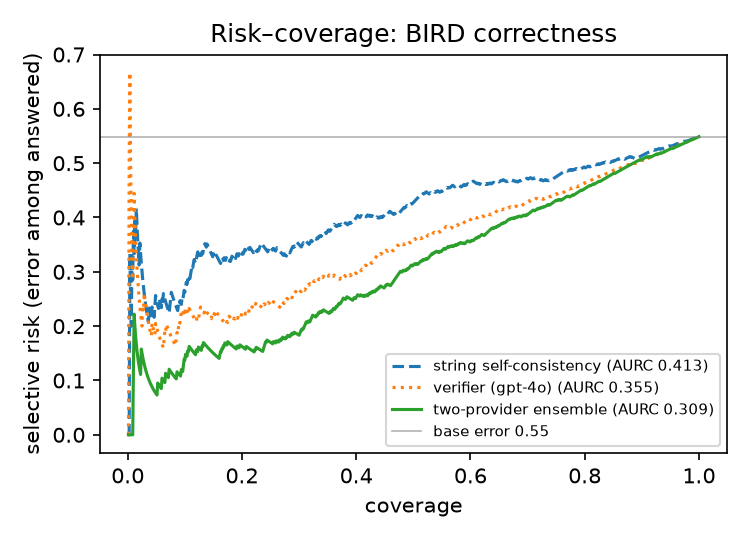}
\hfill
\includegraphics[width=0.48\textwidth]{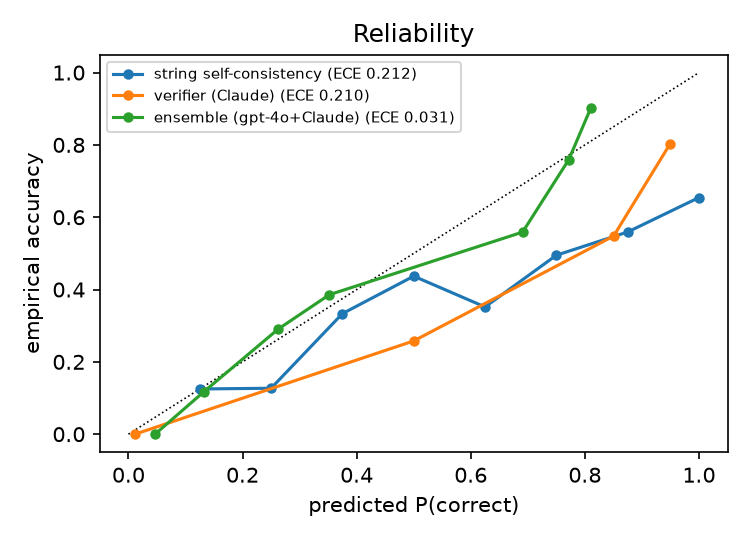}
\caption{Left: risk--coverage, where the ensemble dominates self-consistency. Right: reliability, where the cross-fit ensemble is calibrated (ECE $0.031$) while the raw signals are over-confident.}
\label{fig:rc}
\label{fig:reliab}
\end{figure}

\subsection{Where the errors are: computation, not schema linking}
\label{sec:errors}

The preceding results show that verification works and schema relevance does not. To understand why, we parse each gold query into structural features and examine where the generator fails. The pattern is clear: the difficult cases are not mainly the ones with more joins. They are the ones that require computation and composition.

\begin{table}[t]
\centering
\begin{tabular}{lrrrrrr}
\hline
Gold-query feature & $n$ & Acc. & Drop & SC \auroc{} & Verifier \auroc{} & $\Delta$ \\
\hline
Arithmetic (ratios, \%) & 137 & 0.277 & $0.210$ & 0.716 & 0.813 & $+0.097$ \\
Nested subquery & 50 & 0.280 & $0.183$ & 0.804 & 0.819 & $+0.016$ \\
CASE expression & 63 & 0.286 & $0.180$ & 0.666 & 0.728 & $+0.062$ \\
GROUP BY & 69 & 0.290 & $0.177$ & 0.651 & 0.609 & $-0.042$ \\
DISTINCT & 123 & 0.317 & $0.159$ & 0.584 & 0.639 & $+0.054$ \\
ORDER BY & 163 & 0.331 & $0.151$ & 0.645 & 0.678 & $+0.033$ \\
Multi-table / join & 636 & 0.441 & $0.049$ & 0.674 & 0.779 & $+0.104$ \\
Aggregate & 350 & 0.451 & $\approx 0$ & 0.735 & 0.773 & $+0.038$ \\
\hline
\end{tabular}
\caption{Per-feature analysis on the 800-question slice (features can co-occur). Columns 2--4: modal-query accuracy when the feature is present and the drop versus absent. Columns 5--7: string self-consistency versus the GPT-4o verifier as correctness predictors within that subset. Errors concentrate in computation and composition, and the verifier improves over self-consistency on every feature except GROUP BY.}
\label{tab:error-features}
\end{table}

Join count barely changes accuracy: queries with 0, 1, 2, and 3 or more joins have accuracies $0.48$, $0.45$, $0.41$, and $0.45$, respectively. A logistic model over the same features ranks arithmetic, ORDER BY, and DISTINCT as the strongest independent error predictors, while join and aggregate indicators have negative coefficients. The benchmark's own difficulty labels agree with this reading: accuracy is $0.544$ on simple questions, $0.332$ on moderate questions, and $0.287$ on challenging questions.

This explains why schema relevance is a weak correctness signal. The generator often identifies the relevant tables and columns, but fails on ratios, conditional logic, grouping, ranking, or nested structure. A schema-relevance score cannot see those failures. A verifier can, because it evaluates whether the computation expressed by the SQL matches the question.

To see whether the verifier earns its advantage where the errors actually are, columns 5--7 of Table~\ref{tab:error-features} compare string self-consistency and the GPT-4o verifier as correctness predictors within each feature subset. The verifier improves on self-consistency for almost every feature, and its strongest discrimination is on arithmetic questions (\auroc{} $0.813$), the largest error category. The clearest exception is GROUP BY, the one subset where the verifier ($0.609$) does not beat self-consistency ($0.651$). Subqueries are a near-tie ($+0.016$), but only because self-consistency is already strong there ($0.804$): the model's samples disagree visibly when a query needs nesting, so sampling agreement already flags those failures.

The verifier's advantage is not confined to the hardest queries. Splitting the slice into logic-heavy queries (arithmetic, subquery, CASE, or GROUP BY; $n=225$, accuracy $0.293$) and the rest (simple; $n=575$, accuracy $0.513$), the verifier's edge over self-consistency is actually larger on the simple queries ($+0.117$, from $0.638$ to $0.755$) than on the logic-heavy ones ($+0.073$, from $0.690$ to $0.763$). The reason is that self-consistency is weakest precisely on the simple queries, where the model is often confidently and unanimously wrong: its samples agree, so sampling agreement reports high confidence on a query that is in fact incorrect. The verifier catches these cases because it reasons about the computation rather than counting agreement.

\section{Discussion}

The central lesson is that correctness uncertainty for text-to-SQL is real, but it does not come from the signals that are cheapest to compute. Sampling agreement, structural agreement, execution-result agreement, schema relevance, executability, and even sequence log-probability all occupy the same broad performance band on hard multi-table data. These signals are useful diagnostics, but they mostly measure whether the generator is internally consistent, not whether the generated SQL actually answers the question. In this setting, a model can be highly self-consistent and still be consistently wrong.

Verification changes the picture because it evaluates the candidate query at the level where many errors occur. The difficult cases in our study are not primarily failures to mention the right table or to produce executable SQL. They are failures of computation and composition: ratios, percentages, conditional logic, grouping, ranking, nesting, and distinctness. A confidence signal that only sees agreement among samples has little access to those errors. A verifier, by contrast, can compare the question to the computation expressed by the SQL. This explains why verifier scores move past the black-box ceiling while schema-relevance and self-consistency do not.

The results suggest two different deployment regimes. In a fixed deployment, where the system owner has historical questions and execution labels from the same schemas, a trained verifier is attractive. It is cheap at inference time and performs well in-domain, giving useful abstention frontiers without requiring an additional large-model call. In an open or cross-schema setting, however, fine-tuned verifiers are much less reliable. Across the transfer experiments we tried, small and medium trained models improved in-domain performance but did not close the gap to a large frozen judge on unseen schemas. For that setting, the evidence points toward using a strong reasoning model as the verifier, and toward combining independent-provider judges when the cost is justified.

The multi-provider result is practically important. The strongest single verifier is already well above the black-box baselines, but the best score comes from combining GPT-4o and Claude. Their errors are only moderately correlated, and the calibrated ensemble improves both discrimination and reliability. This suggests that verification uncertainty is not merely a property of one model family. Independent judges appear to notice different failures, which makes the ensemble a better trust signal than either judge alone.

The abstention results should be read as empirical operating points rather than as a claim that the problem is solved. The calibrated ensemble gives useful risk--coverage frontiers, while self-consistency cannot form a low-risk subset on this split. The distribution-free certificate is more conservative, firing only at the loosest target. This is not surprising in a regime where the base generator is correct less than half the time. A stronger generator, more calibration data, or both should make certified coverage more useful. The present result is that a logic-aware score creates a usable empirical frontier and a conservative certified lower bound; sampling agreement does not.

Finally, the paper separates two questions that are often conflated. One question is whether a critic can improve a query by driving a self-correction loop. Another is whether a final query, after whatever generation or correction process produced it, should be trusted. Our self-correction baseline shows why these are different. One reflection round barely improves accuracy and produces severely overconfident self-reported confidence. Even systems that use self-correction still need calibrated selective prediction at the end of the pipeline.

\section{Limitations}

The main experiments use two OpenAI generators and two judge providers. This is enough to test whether the pattern survives a stronger generator and an independent verifier, but it is not a complete survey of current text-to-SQL systems. An open-weight generator would be a useful additional check, especially because generation errors may differ across model families.

The verifier experiments also leave open the question of scale. We find that fine-tuned small and medium verifiers perform well in-domain but do not transfer reliably to unseen schemas. Scaling a generative verifier to 7B improves the in-domain result but does not close the transfer gap. This does not rule out the possibility that a larger fine-tuned verifier, or a verifier trained on a much broader collection of schemas and tasks, could generalize better.

Our distribution-free certificates are conservative. This is partly a sample-size issue and partly a base-accuracy issue: when the generator is correct only about 45\% of the time, a very strong confidence score is needed to certify high coverage at low risk. The empirical risk--coverage curves are therefore the more informative deployment summary in the present study, while the certificates should be viewed as conservative lower bounds.

Finally, the study focuses on deciding whether to trust a generated query, not on improving the query. Verifier-guided reranking, verifier-guided regeneration, and interactive clarification are natural next steps. The present results suggest that such systems should not rely on sampling confidence alone; they should include a component that explicitly evaluates the logic of the candidate SQL.

\section{Conclusion}

On hard text-to-SQL, the signals that most directly predict correctness are not sampling agreement, structural agreement, executability, schema relevance, or log-probability. Those signals plateau in a relatively narrow band. Verification moves past that ceiling because it can reason about whether the SQL expresses the computation requested by the question. Independent-provider verifiers make different errors and combine into the strongest and best-calibrated correctness signal we observe, enabling empirical abstention where self-consistency cannot. Trained verifiers are useful in-domain, but cross-schema transfer remains difficult; for open settings, a large frozen reasoning judge is currently the more reliable option. The practical message is simple: if a system needs to decide whether to trust generated SQL, it should pay for a logic-aware verifier, and it should calibrate that verifier for selective prediction.

\appendix
\section{Transfer ablations}
\label{app:transfer}
Each row is an attempt to make a fine-tuned verifier transfer to unseen schemas; none closes the gap
to the frozen reasoning judge ($0.710$ macro, $0.770$ pooled). Schema diversity adds Spider's 20
schemas (28 in all); reasoning distillation trains a 1.5B model on a strong judge's rationales plus
the verdict, versus verdict-only; cross-benchmark trains on one benchmark and tests on the other.
\begin{table}[h]\centering
\begin{tabular}{lr}
\toprule
attempt & transfer \auroc{} \\
\midrule
base fine-tune (8 schemas, LODO) & 0.659 \\
+ schema diversity (28 schemas, LODO) & 0.682 \\
scale generative judge to 7B (LODO) & 0.662 \\
reasoning distillation (verdict-only / +reasoning) & 0.653 / 0.564 \\
cross-benchmark (BIRD$\rightarrow$Spider / Spider$\rightarrow$BIRD) & 0.713 / 0.672 \\
\midrule
frozen GPT-4o judge (macro / pooled) & 0.710 / 0.770 \\
\bottomrule
\end{tabular}
\end{table}

\section{Verifier prompts}
\label{app:prompts}
Every verifier sees the same content, presented in this order: the schema, the question, the evidence
string (when available), and the candidate SQL, followed by the elicitation request. Two elicitation
formats are used. The logit format (the OpenAI judges) asks for a one-word verdict and reads
$P(\text{correct})$ from the first-token YES/NO probabilities; the verbalized format (Claude, which
does not expose token logprobs, and the method-matched OpenAI verbal judge) asks for a numeric
probability and divides it by $100$.

\noindent\textbf{Logit (YES/NO) judge.} System: \textit{``You are a strict SQL reviewer. Decide
whether the candidate SQL CORRECTLY answers the question (right tables, columns, conditions,
aggregation, and result). Answer with exactly one word: YES or NO.''} The candidate block ends with
\textit{``Is it correct? Answer YES or NO.''}

\noindent\textbf{Verbalized (0--100) judge.} System: \textit{``You are a strict SQL reviewer. Decide
whether the candidate SQL correctly answers the question. Respond with ONLY an integer from 0 to 100:
the probability (percent) that it is correct. Output only the number.''} The candidate block ends with
\textit{``Probability (0-100) that the SQL is correct:''} The returned integer divided by $100$ is the
correctness probability.

\bibliographystyle{tmlr}
\bibliography{references}
\end{document}